\documentclass[letterpaper, 10 pt, conference]{ieeeconf}  

\IEEEoverridecommandlockouts                              
\usepackage[utf8]{inputenc}
\usepackage[T1]{fontenc}
\usepackage{algorithmic}
\usepackage{graphicx}
\usepackage{hyperref}
\usepackage{amsmath}
\usepackage{color}
\usepackage{todonotes}
\usepackage{amssymb}
\usepackage{lipsum}
\usepackage{multirow}
\usepackage{array}
\usepackage{float}
\usepackage{svg}
\usepackage{dsfont}
\usepackage{subcaption}
\hypersetup{breaklinks=true}
\usepackage[
style = ieee,
sorting=none,
backend=bibtex]{biblatex}
\usepackage[normalem]{ulem} 

\newcommand{\ICRAedit}[1]{{\color{black}#1}}

\title{\LARGE \bf
Human-Centered Autonomy for UAS Target Search}

\author{Hunter M. Ray$^{*\dagger}$, Zakariya Laouar$^*$, Zachary Sunberg and Nisar Ahmed
\thanks{Authors are with the Ann And H.J. Smead Aerospace Engineering Sciences Department, 429 UCB, University of Colorado Boulder, Boulder CO 80309, USA. Correspondence: {\tt\small   hunter.ray-1@colorado.edu.}
}
\thanks{\hspace{-4.7mm} $^*$ These authors contributed equally to this work}
\thanks{\hspace{-4.7mm} $^\dagger$Rescuer, UAS Supervisor, Boulder Emergency Squad, 3532 Diagonal Highway
Boulder, CO 80301.}%
}

\begin{document}

\maketitle
\thispagestyle{empty}
\pagestyle{empty}

\begin{abstract}
 
Current methods of deploying robots that operate in dynamic, uncertain environments, such as Uncrewed Aerial Systems in search \& rescue missions, require nearly continuous human supervision for vehicle guidance and operation. These methods do not consider high-level mission context resulting in cumbersome manual operation or inefficient exhaustive search patterns. We present a human-centered autonomous framework that infers geospatial mission context through dynamic feature sets, which then guides a probabilistic target search planner. Operators provide a set of diverse inputs, including priority definition, spatial semantic information about ad-hoc geographical areas, and reference waypoints, which are probabilistically fused with geographical database information and condensed into a geospatial distribution representing an operator's preferences over an area. An online, POMDP-based planner, optimized for target searching, is augmented with this reward map to generate an operator-constrained policy. Our results, simulated based on input from five professional rescuers, display effective task mental model alignment, 18\% more victim finds, and 15 times more efficient guidance plans then current operational methods.

\end{abstract}


\section{Introduction}

Uncrewed Aerial Systems (UAS) are revolutionizing public safety as they provide new perspectives and rapid access to first responders in a variety of tasks including search \& rescue (SAR), firefighting, water rescue, and law enforcement \cite{ray_review_2022}. Operation of UAS often requires up to three team members to respectively pilot the vehicle, act as a technical specialist or observer, and coordinate platform integration across a mission. UAS autonomy can enable single-pilot operations with greater operator mobility and situational awareness by substantially reducing high task loads present in these settings. However, the unpredictable nature of public safety incidents puts challenging requirements on an autonomous system's adaptability to changing circumstances.

Autonomous systems must seamlessly fit into teams to be effective, taking unquestioning direction from an operator and rapidly processing novel information, such as new search requirements, to dynamically (re)program their goals. For a UAS involved in SAR, this direction involves the operator defining their preferences about where and how to search specific areas based on their experience and knowledge, which characterizes a portion of their mental model. However, the unstructured nature of these preferences must be condensed into an accessible form for computationally constrained planning and informed execution. 

\begin{figure}[t]
    \centering{
    \includegraphics[width = \linewidth]{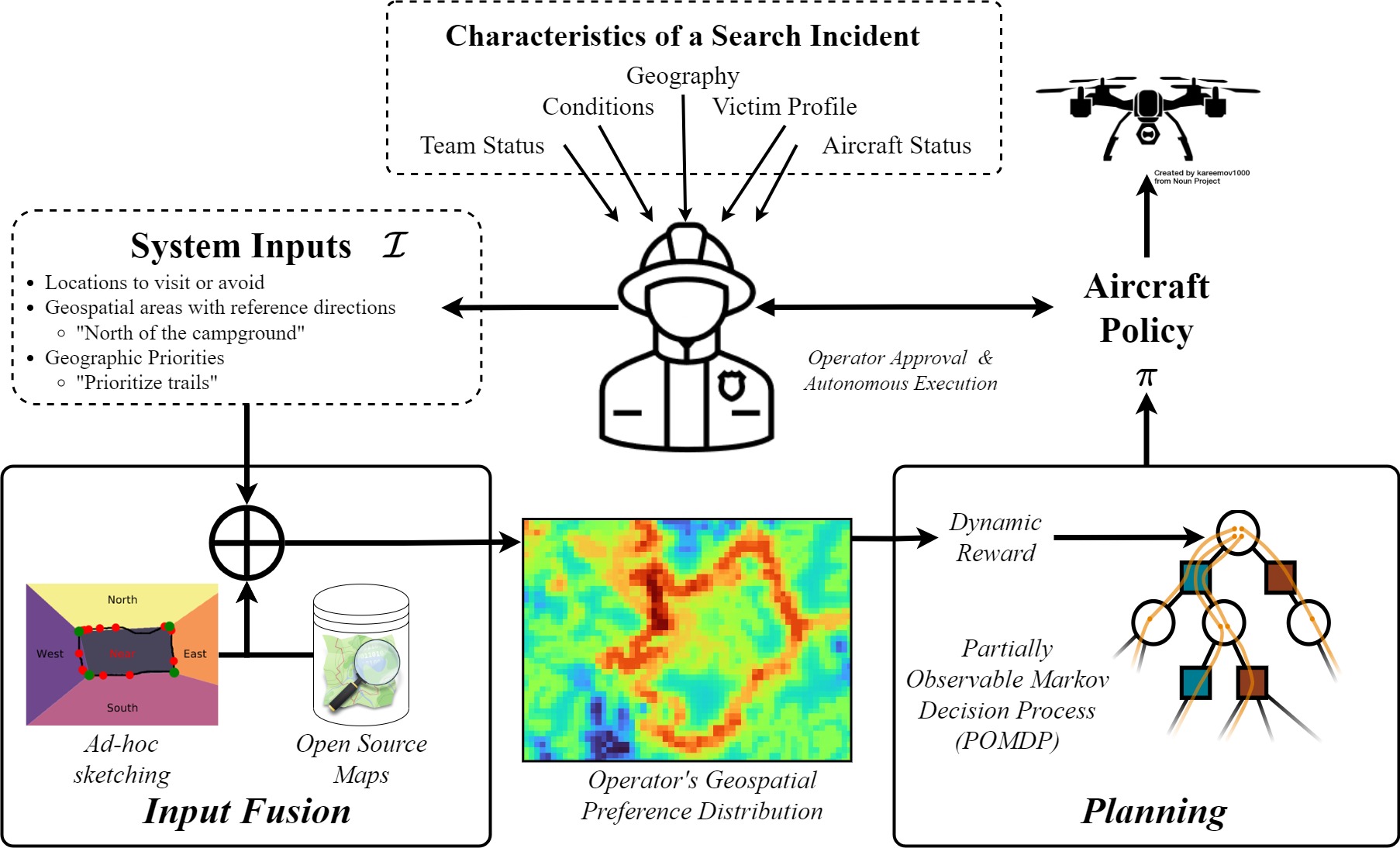}}
    \vspace{-6mm}
\caption{Search \& rescue incidents require operators to fuse multiple sources of information to direct an aircraft. Our human centered architecture fuses a variety of unstructured operator inputs to inform an optimal planning process that generates a policy for autonomous execution.} 
    \label{fig: interaction}
    \vspace{-20pt}
\end{figure}

Building upon our experience in rescue operations, we develop a human-centered autonomous framework, shown in Figure \ref{fig: interaction}, which enables dynamic planning for UAS SAR missions. Current methods of automated flight use an operator's inputs, $\mathcal{I}$, as a deterministic policy, $\pi$, to follow a set of waypoints or execute an exhaustive search over a polygonal area \cite{ray_review_2022}. However, these $\mathcal{I}$ also contain valuable context, which can infer higher level mission goals, and thereby define extensive autonomous behavior. With this new information, the autonomous agent must effectively balance the operator's direction while fulfilling the primary task of searching for the victim.

We build upon current modes of input through an augmented set of $\mathcal{I}$ - priority definition, spatial semantic observations over ad-hoc geographical areas, and example waypoints - which informs an operator's geospatial preferences over a mission area. These inputs are processed using a probabilistic model and fused via Bayesian inference to estimate a geospatial distribution reflecting operator preferences. Our new, model-driven, approach is necessary as no two incidents are exactly the same and attempting to systematically learn overall preferences based on prior events is challenging due to multiple communication modalities, key nuances across incidents, and limited data availability. Instead, we dynamically fuse and then embed operator preferences into a reward model for a Partially Observable Markov Decision Process (POMDP) to generate an online policy. 

In summary, our key technical contributions include: (1) \ICRAedit{a new interaction paradigm} that enables operators to provide diverse modes of input, which inform a nuanced and complex set of preferences on autonomous behavior; (2) a probabilistic model that rapidly aligns task mental models for geospatial feature prioritization between autonomy and expert using sparse training data and operator-augmented feature vectors; (3) a general POMDP formulation for collaborative target search using adaptive reward; and (4) proof-of-concept user study with five rescuers that leverages a realistic search scenario and export informed truth model to evaluate our approach in simulations against an operational baseline.

\section{Problem Statement and Background}
\label{seq: background}
Our work is motivated with a hypothetical incident, derived from real-life scenarios, without loss of generality. In this scenario, a middle-aged man, a frequent fly-fisherman, was reported missing by his wife after he failed to return from a daytime trip to the Walker Ranch Open Space in Boulder County, CO. The victim's vehicle was found at the southern trailhead. The first arriving unit includes a UAS team, which is tasked with performing a preliminary search of the area. 

\subsection{Problem Statement}

A search operation takes place over a larger state space, $S\subseteq \mathbb{R}^f$, representing $f$ mission influencing factors, including static (geography) or dynamic (team locations) variables. The operator acts as a means of fusing the contextual and environmental aspects of an evolving operation. They provide a set of inputs, $\mathcal{I}$, that correspond to their preferences over locations to search. $\mathcal{I}$ must be interpreted along with aircraft performance constraints, such as limited battery life, to generate a guidance policy, $\pi$, that effectively searches an area. The aircraft can perform its internal navigation and control to follow $\pi$ while looking for infrared hotpots or visual clues as to a static victim's location. This detection can rely on computer vision or the operator to supervise video feeds, however this is outside this paper's scope.




\subsection{Related Work}
Given the information in the motivating mission, an officer directing operations gains awareness of the victim's physical condition, potential equipment, last-know-point, mindset, and behavior. These factors of their evolving mental model are then grounded in the mission area, which informs the resulting search methods and tasking \cite{rouse_looking_1986}. If the UAS management system can minimize the need for input and monitoring, the operator can move across the environment and maintain situational awareness, resulting in a faster rescue time. However, this requires interpreting and then incorporating the mental model into the overall autonomy architecture.

While robots interpret mental models in multiple ways \cite{tabrez_survey_2020} and teams that use them improve their performance \cite{gervits_toward_2020}, they often rely on static, pre-defined architectures \cite{scheutz_framework_2017},\cite{albrecht_autonomous_2018}. Autonomous behavior has also been effectively guided by operators using a variety of methods, including implementing a POMDP to couple the operator's inputs with the search task planning \cite{jamieson_active_2020}, leveraging set plays \cite{miller_delegation_2014}, or using Bayesian priors on geological knowledge to define operator preferences \cite{arora_multi-modal_2019}. While these methods effectively leverage the operator's knowledge to guide behavior, they require limited, specialized, or mission specific inputs. We aim to decouple the fusion of the operator's inputs with the vehicle planning and execution to enable a more flexible, `plug-and-play', approach that accommodates diverse underlying autonomous architectures, such as interchangeable input types, fusion methods, or planners.

An operator's preferences have also been learned using Inverse Reinforcement Learning (IRL). However, whereas IRL relies on observed expected behaviour, such as near optimal reference trajectories, to directly infer reward functions \cite{arora_survey_2021}, we infer higher-level preference distributions by fusing multi-modal inputs and readily available geographic database information. This distribution is factored into a reward function, complementing a planner's baseline performance with end user expertise, addressing the challenges found with rewards designed by engineers \cite{hadfield-menell_inverse_2017},\cite{booth_perils_2023}. 

Leveraging autonomy to aid rescuers in efficiently searching an area provides obvious benefits. Recent work in automated UAS flight planning \cite{kingston2016automated} provides effective search patterns (e.g. lawnmower) often used by U.S. maritime SAR personnel \cite{thomas2013us}. While this can be improved with more explicit modeling of target location uncertainty \cite{bourgault2006optimal}, obtaining an accurate and informed prior is challenging. The fusion of our inputs into a preference distribution could be applied as a target location estimate, however, we aim to preserve mission level uncertainty of target location while accounting for dynamic operator preferences. 

POMDPs are effective at modeling these competing objectives (finding the victim and satisfying their operator) in their ability to reason over aleatoric and epistemic uncertainty \cite{kochenderfer2022algorithms}. POMDPs have already been applied to SAR, but these explicitly model domain information such as cell signals and crowd reports \cite{baker2016planning}. We don't assume this information is available and instead use a POMDP to model expert-provided domain knowledge as dynamic reward, allowing observations to inform and shape the target belief over time. This allows us to flexibly balance the operator's preferences against the agent's primary goal of finding the victim.


 

Methods for searching cluttered environments, such as kitchens, have applied POMDP planners. These complex spaces have been navigated by condensing them into multiple resolutions \cite{zheng2021multi}, incorporating natural language from the user \cite{wandzel2019multi}, or sequentially decluttering the space \cite{xiao2019online}. While these studies focus on close quarters, we investigate planning over multiple square kilometers and flexibly integrating expressive human input from semantic language and sketches. 



Compared to previous work, we rely on the operator to interpret the vast set of unstructured data involved in a SAR mission, which is conveyed through intuitive methods and fused with geographical data. During execution, our approach allows for \textit{human-on-the-loop} collaboration, where the human can provide inputs at any time during the search, but the system can still function competently without input. This paradigm enables the autonomous system to effectively balance the overarching goal of finding the victim while being informed and guided by dynamic operator preferences.



\section{Methodology}
\label{seq: methods}
A two-step approach is implemented towards interpreting user inputs and generating an aircraft policy, as shown in Figure \ref{fig: interaction}. An operator's inputs are captured and fused via a probabilistic graphical model to define a geospatial preference distribution. This distribution augments a POMDP reward function, generating a dynamic policy to inform aircraft guidance. 

\subsection{Input Fusion}
\begin{figure}[t]
    \centering{
    \includegraphics[width = 0.85\linewidth]{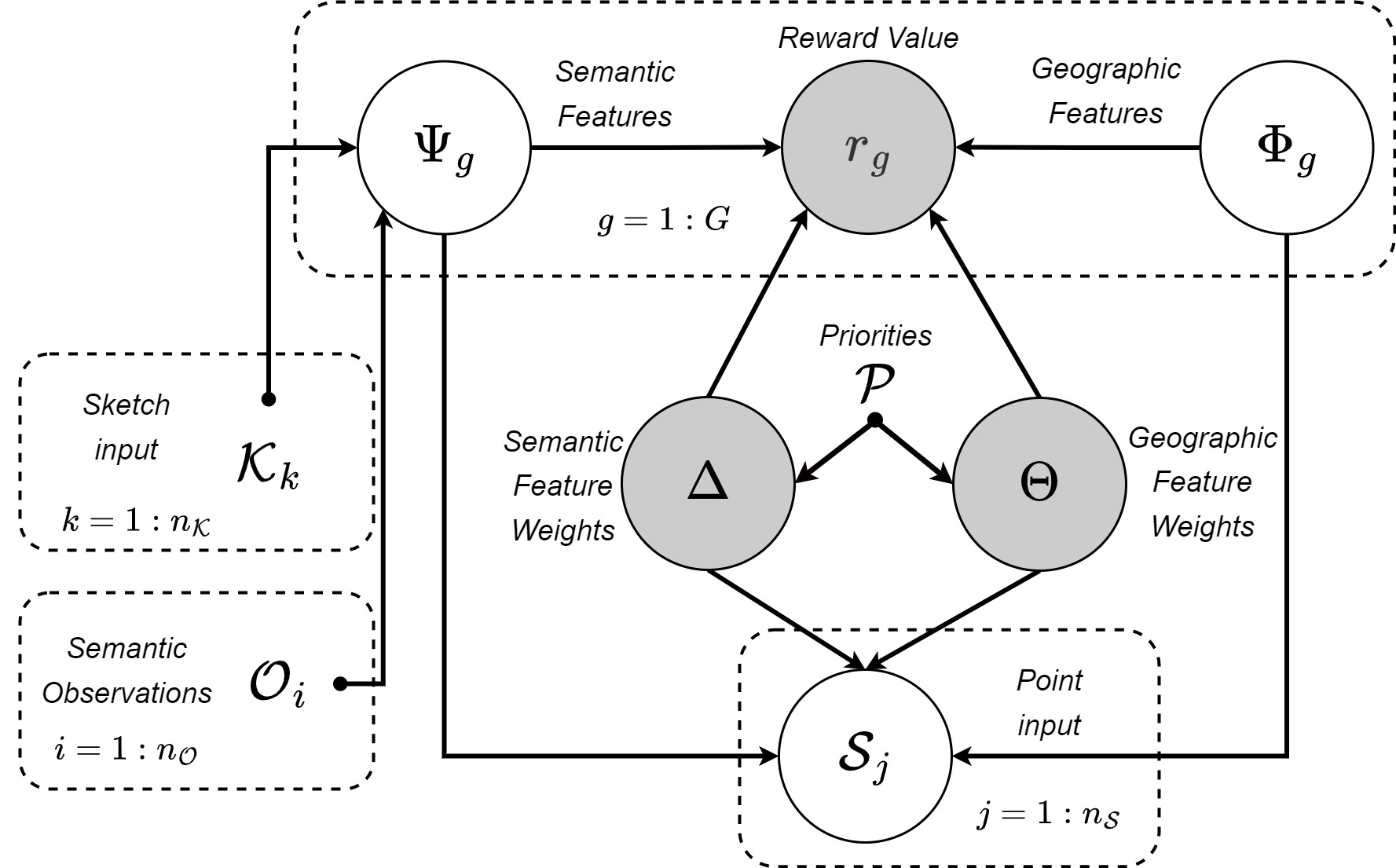}}
    \vspace{-2mm}
    \caption{Graphical model used in \ICRAedit{input fusion} algorithm with observed variables (white) and unobserved variables (grey). }
    \label{fig: graphical_model}
    \vspace{-5mm}
\end{figure}
 We define a grid $\mathcal{G} \subset S$, which discretizes the operational region within $S\subseteq \mathbb{R}^2$ into an $n\times m$ grid, $\mathcal{G}$, with areas $g\in \mathcal{G}$ of uniform resolution. Each $g$ contains a set of defining features compactly represented as vectors. This includes a static set of geographic features, $\Phi \in \mathbb{R}^{n_\Phi}$, which describe how $g$ is related to the built and natural environment including trails or buildings. Each $g$ is augmented with a set of user-defined features, $\Psi\in \mathbb{R}^{n_\mathcal{O}}$, which describe $g$'s spatial context within a particular mission, such as its proximity to a staging area or location within a specific map sector. 
 
 We assume that the operator's preferences can be modeled as a set of visitation preferences over $\Phi$ and $\Psi$, which are represented as an unnormalized set of associated feature weights for geographic, $\theta \in \mathbb{R}^{n_\Phi}$, and semantic, $\Delta \in\mathbb{R}^{n_\mathcal{O}}$, components. For a given $g$ with operator reward, $r$, the associated set of features, $\Phi_{g}$ and $\Psi_{g}$, are modeled to be linearly related to their respective weights, $\theta$ and $\Delta$, 
 \allowdisplaybreaks
\abovedisplayskip = 2pt
\abovedisplayshortskip = 2pt
\belowdisplayskip = 0pt
\belowdisplayshortskip = 0pt
\begin{align}
    r_g = \theta^T\Phi_g+\Delta^T\Psi_g
    \label{eq: reward}
\end{align}

This linear relation enables the different features to be appropriately accounted for while being flexible to changing $n_{\mathcal{O}}$. To account for uncertainty from imperfect mapping, limited or mistaken operator input and other unknown error sources, we seek to infer a probability distribution over $r_g$, $p(r_g|\mathcal{I})$. This inference is accomplished through the graphical model shown in Figure \ref{fig: graphical_model}, which the operator engages with through a set of three possible inputs.

We define the set of $\mathcal{I}$ to include $n_\mathcal{P}$ priorities $\mathcal{P}$, $n_\mathcal{O}$ semantic geospatial observations $\mathcal{O}$, and $n_\mathcal{S}$ waypoints, $\mathcal{S}$. All three types of $\mathcal{I}$ can be provided at the start and while it is amenable to modification during the mission this is not explicitly shown in our experiments. Each $\mathcal{P}$ acts as a positive, equally weighted prior on the respective geographic or semantic feature's weight. An input of $\mathcal{O}$ is built with a combination of a 2D convex sketch, $\mathcal{K}$, and a geospatial semantic reference label, where each additional $\mathcal{O}$ increases the dimensions of $\Delta$ and $\Psi$. Finally, the set of $\mathcal{S}$ includes locations for the aircraft to visit $(S=1)$ and those that should be avoided $(S=0)$. 

Within the context of our reference mission, these inputs may take the following form: $\mathcal{P}=\left[\mathrm{Trails},\mathrm{\;Section\: A}\right]$, $\mathcal{O} =$ ["Search north of the Bridge", "Search inside section A"], $\mathcal{K} = \left[\mathrm{Bridge}, \mathrm{\;Section\: A}\right]$, $\mathcal{S}_{\mathrm{visit,\:}(S=1)} = $ [Set of waypoints along a river], $\mathcal{S}_{\mathrm{avoid,\:}(S=0)}=$ [Set of waypoints near structures].

While the set of $\mathcal{P}$ and $\mathcal{O}$ \ICRAedit{initialize} the estimates and add features to the environment, the observable set of $\mathcal{S}$ help tie the known feature components $\Phi$ and $\Psi$ to their respective unknown weightings $\theta$ and $\Delta$. We assume that each $g$ must reach a specific threshold of an operator's optimal positive or negative preference for it to be provided as a reference. We therefore model $p(S_g|\theta,\Delta,\Phi_g,\Psi_g)$ as a logistic function in equations \ref{eq: points_pos} and \ref{eq: points_neg} with $r_g$ defined as in equation \ref{eq: reward}, 
\begin{align}
    p(S_g=1|\theta,\Delta,\Phi_g,\Psi_g) &= \frac{\exp(r_g)}{1+\exp(r_g)},
    \label{eq: points_pos}
    \\
    p(S_g=0|\theta,\Delta,\Phi_g,\Psi_g) &= \frac{1}{1+\exp(r_g)}.
    \label{eq: points_neg}
\end{align}

To define $\Phi$, geographic features can be extracted using publicly available government and open source datasets from across the world. A geographical information system software, such as ArcGIS, can integrate a \ICRAedit{subset of information for a specific area, such as a county or an agency's jurisdiction. For this work, we consolidate information on} 
roads, trails, structures, stream lines, water bodies, and tree canopy. We aim to create a set of $\Phi_g$ that accurately maps an operator's perspective of their respective value, which includes each $g$'s proximity to a relevant feature. Therefore for each $g$ with a given resolution, we determine the distance, $d$, to the closest respective feature, $f$, informing an adjacency metric, calculated as $\Phi_{f,g} = \exp(\frac{-d}{\textit{resolution}})$. The geoprocessing can be performed offline and saved into a resolution specific database.

To define $\Psi$, prior work \cite{ahmed_bayesian_2013, burks_harps_2022, sweet_structured_2016, burks_active_2020} is leveraged that probabilistically relates a geospatial semantic label to a given sketch, $\mathcal{K}_k$. To reduce computational loads, we sequentially reduce the number of vertices to five points that maximize the overall area to form the resulting convex polytope. The available semantic labels includes a comprehensive set of canonical bearing labels \{"N","NE","E","SE","S","SW","W","NW"\} and discrete ranges \{"Inside", "Near", "Outside"\}. Given a particular grid area $g$, with a center location $x_g\in \mathbb{R}^2$ we can approximate the likelihood of the semantic label using the softmax function,  where each class contains a set of parameters $w\in \mathbb{R}^2$ and $b\in\mathbb{R}^1$, which constrain their boundaries along the sketch border \cite{sweet_structured_2016}. Once defined, a Monte Carlo approximation is used to correlate the softmax classes with specific semantic labels \cite{burks_active_2020}, resulting in $p(\mathrm{label}|\mathrm{class})$. We therefore define each $\Phi_g$ as $p(\mathrm{label}|g)$, the probability of the given grid point being represented by a certain label,
\begin{align}
    \Psi_{i,g}=p(\mathrm{label}|g) = \sum_{class} p(\mathrm{label}|\mathrm{class})p(\mathrm{class}|g) 
\end{align}

Having defined the specific components of the graphical model, inference must be performed over $p(\theta,\Delta|\Phi,\Psi,\mathcal{P},\mathcal{S})$ to find the resulting reward. The weights, $\theta$ and $\Delta$, are each represented as multivariate Gaussians with respective means $\mu_\theta , \mu_\Delta$ and covariances $\Sigma_\theta , \Sigma_\Delta$, allowing a quick approximation of the distribution through the Laplace approximation \cite{bishop_pattern_2006}. We take the resulting expected value of $\theta$ and $\Delta$ to estimate the posterior reward: 
\vspace{-1pt}
\allowdisplaybreaks
\abovedisplayskip = 2pt
\abovedisplayshortskip = 2pt
\belowdisplayskip = 0pt
\belowdisplayshortskip = 0pt
\begin{align}
    \hat{r}_g &= \mu_\theta^T\Phi_g+\mu_\Delta^T\Psi_g  \label{eq: reward_est}
\end{align}

\noindent Additional derivations and details on model inference are discussed in the extended version of our paper \cite{ray2023humancentered}.

\subsection{Adaptive-Reward Target Search POMDP} \label{sec:pomdpform}
The objective of the planning module is to search for a static target with an unknown location while accounting for operator preferences. We do this by formulating and then solving a general target search POMDP that leverages the operator's geospatial preference distribution to find the target within the previously defined $n\times m$ grid, $\mathcal{G}$. We introduce our POMDP formulation below: 

A \emph{Partially Observable Markov Decision Process} (POMDP) is a tuple $(\mathcal{S}, \mathcal{A}, \mathcal{O}, \mathcal{T}, \mathcal{R}, \mathcal{Z}, \gamma)$, where:
$\mathcal{S}, \mathcal{A},$ and $\mathcal{O}$ are finite sets of states, actions and observations, respectively, 
$\mathcal{T} : \mathcal{S} \times \mathcal{A} \times \mathcal{S} \rightarrow [0,1]$ is the transition probability function,
$\mathcal{R} : \mathcal{S} \times \mathcal{A} \rightarrow \mathbb{R}$ is the immediate reward function,
$\mathcal{Z} : \mathcal{S} \times \mathcal{A} \times \mathcal{O} \rightarrow [0,1]$ is the probabilistic observation function,
$\gamma \in [0,1)$ is the discount factor \cite{kochenderfer2022algorithms}.

\textbf{State space} $\mathcal{S}$: Four components: 2D robot position, 2D target position, and battery life of the search agent. Lastly, the state is augmented with a $|n\times m|$ bitvector to encode whether a robot has visited each grid cell.

\textbf{Observations} $\mathcal{O}$: The target location uncertainty is represented as a one-hot vector encoding of target occupancy in the $\mathcal{M}$ cells adjacent to the true target cell. Additionally, the agent may receive a \textit{target-not-found} observation represented as a vector of zeros, leading to a cardinality of $|\mathcal{M} + 1|$. 

\begin{figure}[t!]
  \centering
    \includegraphics[width=0.8\linewidth]{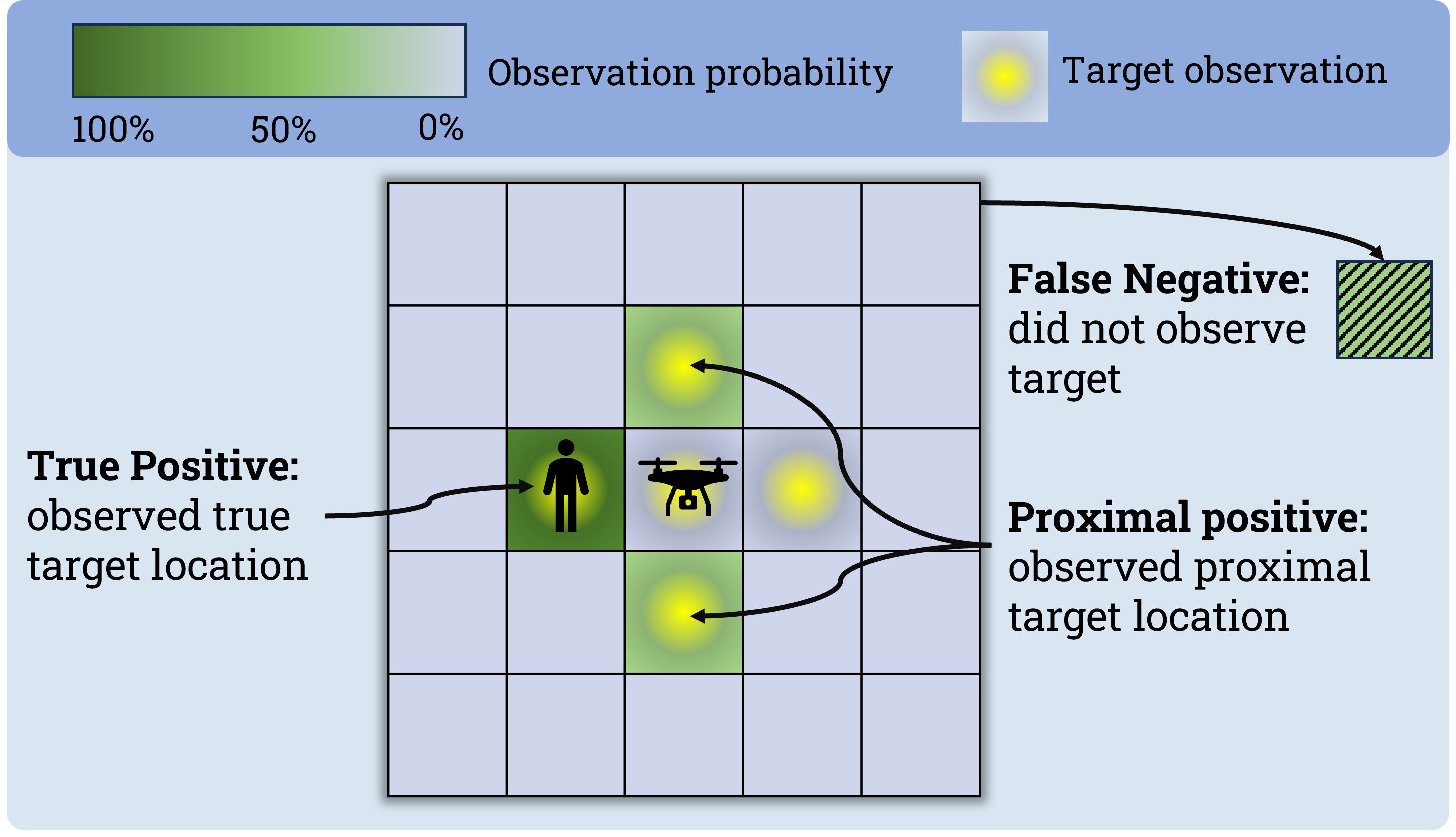}
    \vspace{-2mm}
    \caption{POMDP observation model.}   
    \label{fig:obs}
    \vspace{-6mm}
\end{figure}

The observation function, $\mathcal{Z}$, models the agent's ability to accurately observe the target, informed by factors such as sensor capabilities or terrain complexity. This nuance can be expressed by choosing a Manhattan distance, $D_{obs}$, to express how close the agent must be to receive a positive target observation. The agent receives a noisy observation of the target at each timestep. If within $D_{obs}$, the agent can receive one of the following three observations given the true location of the target. The agent can receive a true positive observation with probability $Z_{true}$. The agent can also receive a \textit{proximal observation} which returns a target observation near the true target location. Each cell defined as proximal to the true target cell will receive this observation with probability $Z_{prox}$. Lastly, the agent can receive a false negative observation with probability $Z_{neg}$ as shown in Figure \ref{fig:obs}. Otherwise, the agent receives the \textit{target-not-found} observation with 100\% probability.

In the example shown in Figure \ref{fig:obs}, the agent receives a positive observation only if adjacent to the true target. Thus, $\mathcal{O}$ is presented with $\mathcal{M}=5$ and $\mathcal{Z}$ being distributed over the true target cell and the 2 adjacent cells. 

\begin{figure*}[t]
    \centering{
    \includegraphics[width = \linewidth]{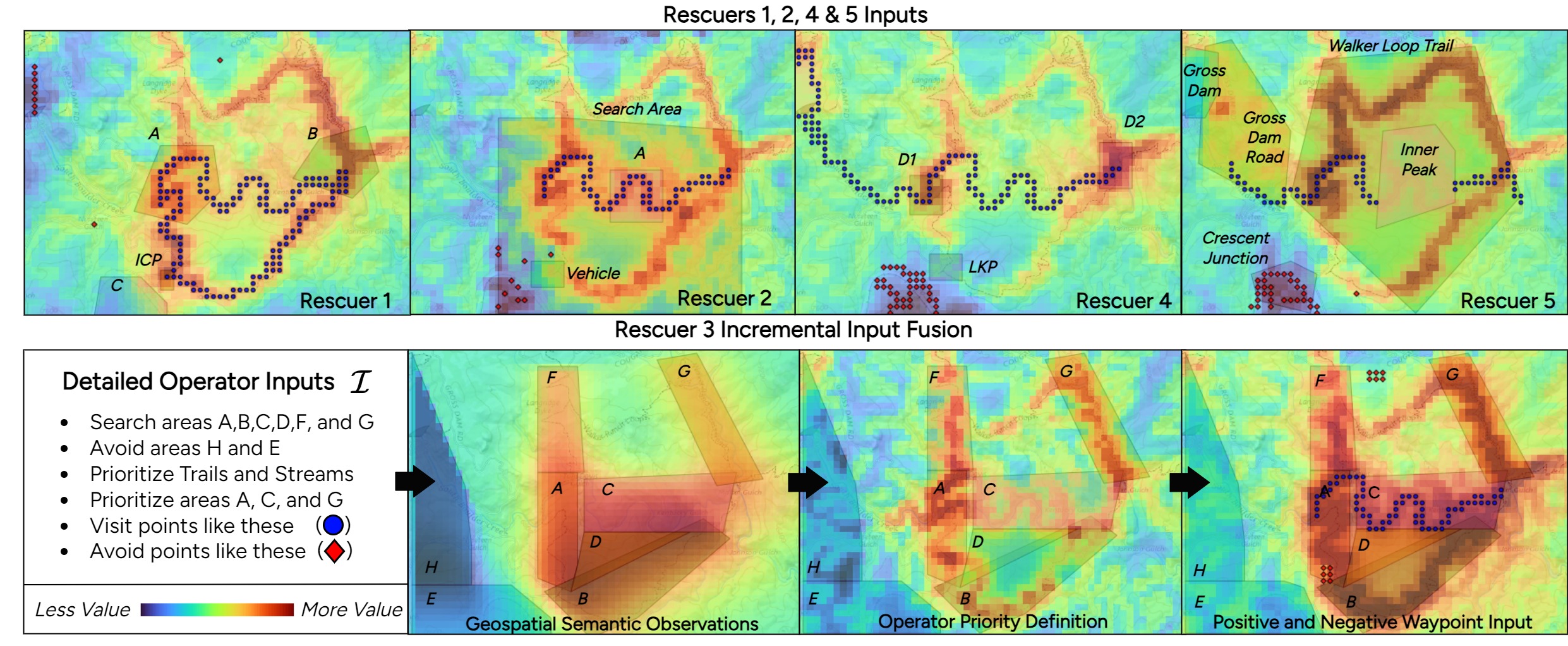}}
    \vspace{-6mm}
    \caption{Inputs from five rescuers overlay the resulting geospatial preference distribution. The lower row shows the progressive addition of rescuer 3's inputs and resulting concentration of reward around trails and streams within prioritized areas. \ICRAedit{Note that in execution, inputs are fused simultaneously.} }
    \label{fig: operator_data}
    \vspace{-5mm}
\end{figure*}

\textbf{Transitions} $\mathcal{T}$: The agent can deterministically move in four cardinal directions: \textit{up, down, left, right} as well as  \textit{stay} in the grid cell. As the target is assumed to be static, their position remains the same. The state's battery component decreases a fixed $B_{cost}$ amount each timestep. When the agent transitions from state $s$ to $s'$, the occupancy history is updated, ensuring a record of $g$ that the agent has visited. 

\textbf{Rewards} $\mathcal{R}(s,a,s')=R_{time} + R_{target} + R_{op}$
\begin{itemize}
    \item $R_{time}$: Time penalty for each step taken.
    \item $R_{target}$: Reward collected for finding the target, i.e. if the agent and target positions in $s'$ are equivalent.
    \item $R_{op}$: Each grid cell $g$ has a corresponding reward $\hat{r}_g$ obtained from the operator's geospatial preference distribution. To incorporate operator preferences, $R_{op}$ will only be collected if the robot location in $s'$ has yet to be visited: $R_{op}=\hat{r}_g \times \mathds{1}_{visited(s')}$.
\end{itemize}

To incentivize the agent to travel to the regions that have a concentrated belief over the target location, the reward for finding the target needs to be higher than the highest value in the operator's geospatial preference distribution. If this distribution provides higher reward regions, the agent will more likely seek out those regions, ignoring valuable information about the target's potential location.

\textbf{Termination:} A \textit{terminal} or \textit{absorbing} state is a state from which no further transitions to other states are possible. Conditions for termination are: (i) the current robot position equals the target position, and (ii) the current battery equals the battery required to return to the initial state, with the initial battery life $B_{max}$ tuned to represent operational constraints. Modeling the battery life as a terminal state ensures that the aircraft can return to base before the battery depletes and, in doing so, incentivizes exploration closer to base.


\subsection{Planning}
Our POMDP formulation results in a very large state space ($|\mathcal{S}| = B_{max} \times (n \times m)^2 \times 2^{(n \times m)}$) to plan over. As SAR requires fast and efficient planning, offline POMDP solvers are too slow for practical application. 
\ICRAedit{While amenable to any online Monte Carlo based solver, leverage POMCP~\cite{silver2010monte} and capture the belief over target location using a particle filter to overcome scalability challenges associated with planning in high dimensional spaces.}

\textbf{Rollout Policy:} A simple and effective rollout policy is a practical method for evaluating nodes of an online POMDP solver. Our rollout policy relies on an MDP abstraction of the larger POMDP by removing observations and only considering the agent's position, battery, and a randomly sampled target position. We solve the abstraction using discrete value iteration upon problem initialization. During the execution of the POMDP, this policy is evaluated for candidate methods of exploring the state space and, once completed, collects the full reward including the operator's preference distribution. While not accounting for agent's current belief or operator preferences, this approach retains acceptable grid exploration with low computational cost.


\section{User Evaluation Results and Discussion}
\label{seq: results}

In this section, our approach is validated against an operational baseline in a realistic simulation environment. Results show that \ICRAedit{without loss of generality with respect to operator input}, our approach can reliably interpret and plan over an operator's inputs to more efficiently find a missing person than current operational standards for autonomous flight. 

To that end, inputs from five first responders were collected from the Boulder Emergency Squad (BES), a volunteer technical rescue team in Boulder, CO that has been using UAS for over eight years in SAR, firefighting, and law enforcement operations \cite{ray_review_2022}. Subjects were ages 24-61 and were either UAS pilots or had supplementary accreditation in search management, reflecting subject matter expertise. Each rescuer provided a set of inputs, $\mathcal{I}$, based on the Lost Fisherman scenario discussed previously. Their inputs informed randomized simulations with multiple vehicle launch locations and target positions, whereby we compare our system's performance against an operational baseline.

\subsection{Data Collection}
To collect subject data, the researcher first walked subjects through a testing scenario detailing the inputs they could provide. Subjects were guided towards appropriate inputs when necessary, such as using the priority definition when they wanted certain features to be focused on. To replicate an intuitive interaction method uninhibited by software limitations, inputs were collected on a paper map, which was then translated into the reward model and inference routine post-hoc. Figure \ref{fig: operator_data} shows\ICRAedit{, for illustration,} the progressive fusion of rescuer 3's inputs and the final fusion results for rescuers 1, 2, 4 and 5. All rescuers provided unique inputs, although often promoted similar features, such as trail intersections and primary streams. In the lower part of Figure \ref{fig: operator_data}, \ICRAedit{rescuer 3's inputs build the reward map} with their preferences concentrating in the prioritized areas and where the trail travels close to streams.

\subsection{Input Fusion Validation}
Before evaluating the overall framework, accurate fusion of an operator's inputs must be validated by aligning the algorithm's estimate with their true geographic and feature preferences. This is accomplished with an error metric that compares the uncertainty-weighted reward estimate of 21 distributed locations against an operator's quartile ranking. In addition, the algorithm's positive and negative feature prioritization is computed to understand the underlying motivation for alignment. This evaluation adapts a search engine ranking metric known as normalized Discounted Cumulative Reward \cite{jarvelin_cumulated_2002}, which normalizes the estimated ranking order by the subject-provided ranking to provide an optimal value of 1. This metric is repetitively evaluated with a set of weights sampled from $p(\theta,\Delta|\Phi,\Psi,\mathcal{P},\mathcal{S})$. Additional details on these metrics is included in the extended version of our paper \cite{ray2023humancentered}.

Input validation results are shown in Table \ref{tbl: RINAO Alignment}, where each metric is compared to a random baseline, shown in parentheses, which is used as no comparable baseline exists. Results across all subjects show very low error values. Positive feature rankings mostly show good alignment, though alignment rank suffered if limited inputs were provided or subjects provided feature weightings that were close together, such as for Rescuer 4. Negative feature rankings proved to be more effective, thereby highlighting that the system can accurately discriminate features that should be avoided.

\begin{table}[t]
\begin{center}
\caption{\label{tbl: RINAO Alignment}Alignment results (with random baseline)}
    \begin{tabular}{{  c|c|c|c  }}
         \hline
         \multirow{2}{3em}{Rescuer} & Error & Pos. Feature Rank & Neg. Feature Rank  \\ 
           & (Optimal = 0) & (Optimal = 1) & (Optimal = 1)\\[0.5ex] 
         
         \hline 
         1 & \textbf{0.015} (2.95) & \textbf{0.93} (0.63) & \textbf{1.21} (3.66)  \\  
         2 & \textbf{0.015} (2.19) & \textbf{0.97} (0.89) & \textbf{1.07} (1.21) \\
         3 & \textbf{0.013} (2.41) & \textbf{0.95} (0.72) & \textbf{1.28} (2.17)  \\
         4 & \textbf{0.025} (2.41) & \textbf{0.84} (0.55) & \textbf{1.85} (2.02)  \\
         5 & \textbf{0.009} (2.06) & \textbf{0.94} (0.76) & \textbf{1.47} (1.81)  \\  
         \hline
         
    \end{tabular}
\end{center}
\vspace{-25pt}
\end{table}

\subsection{Search and Rescue Simulation}

Based on the value maps generated from the rescuers' inputs, the planner is simulated and its performance is compared to operational methods. The environment consists of a 44 $\times$ 59 grid, with each grid cell spanning 100 $\times$ 100 meters. Simulations are initialized across four expert-informed start locations and sample target locations from a truth model, shown in our extended version \cite{ray2023humancentered}, which was informed by BES Chief Andy Amalfitano, an expert in SAR. 

\textbf{Baseline Implementation:} The baseline search algorithm 
receives the operator's sketch set $\mathcal{K}$, positive ("Inside" and "Near") observations, $\mathcal{O}$,  and waypoints $\mathcal{S}_{visit}$. \ICRAedit{ This approach reflects currently available autonomous modes used in operations \cite{ray_review_2022}.} The search agent starts at one of four locations and follows a shortest path algorithm to traverse all waypoints. Once complete, the agent travels to each sketched polygon and executes a lawnmower search pattern of the space. If the target is still not found, the agent redirects to the exhaustive search stage and conducts a lawnmower search over the entire environment.


\textbf{Our Approach:} The POMDP described in Sec. \ref{sec:pomdpform} is solved with the following details. We define a strict target observation condition $D_{obs}=1$, defining positive observations of the target only when within $1$ grid cell. If within $D_{obs}$, the UAS receives a true positive observation with $Z_{true}=80\%$ and a proximal positive observation with $Z_{prox}=10\%$ for the two adjacent cells in the direction of the target. We set our reward model's running cost $R_{time}=-1$ and the reward for finding the target $R_{target}=1000$, the $\hat{r}_{g}$ leveraged from the geospatial preference distribution. Finally, we set the UAS initial battery $B_{max}=1000$. 

\begin{figure}[t]
    \centering
    \begin{subfigure}[b]{0.48\linewidth}
        \centering
        \includegraphics[width=\textwidth]{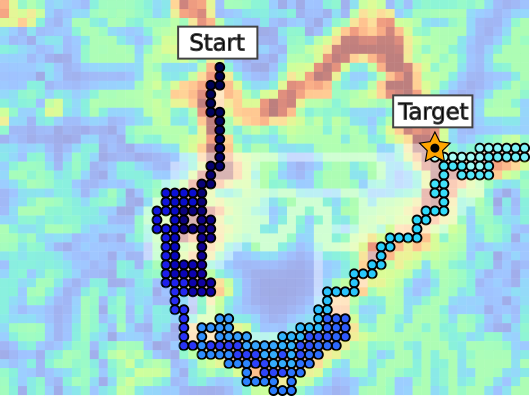}
        \caption{Our approach planning over the underlying geospatial preference distribution.}
        \label{fig:sub1}
    \end{subfigure}
    \hfill
    \begin{subfigure}[b]{0.48\linewidth}
        \centering
        \includegraphics[width=\textwidth]{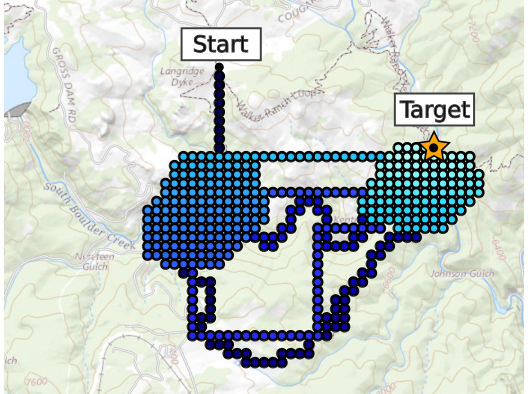}
        \caption{Baseline approach planning over raw operator inputs.}
        \label{fig:sub2}
    \end{subfigure}

    \caption{Example trajectories of our planner and the baseline simulated using rescuer 1 input data from Figure \ref{fig: operator_data}. Our approach finds the target in 329 timesteps, baseline finds the target in 858 timesteps. Trajectory opacity decreases with search time.}
    
    \vspace{-5mm}
    \label{fig: hippo_behavior}
\end{figure}

The resulting UAS behavior is represented in Figure \ref{fig: hippo_behavior}. Our approach efficiently explores high value areas, autonomously searching in and outside of the defined sectors and waypoints, eventually finding the target 2.6 times faster. Our approach is comprehensively evaluated against the baseline using two metrics. The first, \textit{Localization Ratio} defines the ratio of simulation runs that find the target over runs that do not find the target within the given battery life. The second metric, \textit{Reward/Timestep} expresses the discounted reward accumulated per timestep. If the agent fails to find the target under the battery constraints, the simulation is terminated. For each of the four start locations, 100 simulations are performed with the overall reported mean and standard error of the mean reported in Table \ref{tbl: sim_results}.

Our method results in statistically significant $(p<0.05)$ improvement from the baseline approach for all rescuers' input and metrics. Evaluating the localization ratio, our method finds the target 18\% more than the baseline within the available battery life. A two-tailed Binomial test ($N=400$) is performed to find statistical significance. More importantly, the system collects 15.4 times more reward per timestep, reflecting efficient path planning that follows the operator's geospatial context drawn from a limited set of inputs. Statistical significance for this metric is evaluated using a two sample Z-test ($N=400$).

\begin{table}[h]
\vspace{-3mm}
\begin{center}
\caption{\label{tbl: sim_results} Simulation Results - 100 runs per start position}
    \begin{tabular}{{ c|c c|c c }}
         \hline
         \multirow{2}{3em}{Rescuer} & \multicolumn{2}{c}{Localization Ratio}\vline & \multicolumn{2}{c}{Reward/Timestep} \\ 
         
          & Ours & Base  & Ours & Base\\
         \hline
         1 & \textbf{75.3\%} & 54.8\% & \textbf{4.56 $\pm$ 1.21} & 0.490 $\pm$ 0.043\\
         2 & \textbf{54.5\%} & 23.3\% & \textbf{2.05 $\pm$ 0.60} & 0.195 $\pm$ 0.027\\
         3 & \textbf{63.0\%} & 46.3\% & \textbf{4.72 $\pm$ 0.97} & 0.173 $\pm$ 0.023\\
         4 & \textbf{44.3\%} & 25.5\% & \textbf{4.70 $\pm$ 1.36} & 0.101 $\pm$ 0.015\\
         5 & \textbf{51.0\%} & 45.8\% & \textbf{1.85 $\pm$ 0.72} & 0.201 $\pm$ 0.027\\
         \hline
         Average & \textbf{57.6\%} & 39.1\% & \textbf{3.58 $\pm$ 0.67} & 0.232 $\pm$ 0.067\\
         \hline
    \end{tabular}
\end{center}
\vspace{-17pt}
\end{table}

\section{Conclusion}
\label{seq: conclusion}
We introduced a human-centered autonomy framework tailored to UAS deployed in dynamic and uncertain environments, particularly in SAR missions. Our approach hinges on inferring geospatial context and operator preferences using minimal operator inputs to guide a probabilistic target search policy. An operator defines priorities, spatial semantic observations over ad-hoc geographical areas, and waypoints as inputs, which provide valuable context of the operator's mental model. The system infers a reward function from these inputs and plans online using this reward signal and its own belief about target location. Effective alignment of inferred and true operator preferences is proved through a spatial error metric and feature ranking. Simulated Monte Carlo trials revealed that our inference and planning pipeline significantly outperforms an operational baseline in both target localization and reward accumulation. 

While our application focuses on SAR, the ability to estimate an operator’s geospatial preferences for guided planning shows promise in enabling expert-directed autonomy in any data-driven environment. However, our method is limited to areas with mapped geography. Future work will seek ways to address this limitation, such as incorporating online aerial perception techniques, and deploying fielded hardware experiments with vehicle and interface integration.

\section*{Acknowledgements}
We thank the members of the Boulder Emergency Squad for their thoughtful contributions and input.

\printbibliography

@phdthesis{burks_active_2020,
	address = {Boulder, CO},
	title = {Active {Collaborative} {Planning} and {Sensing} in {Human}-{Robot} {Teams}},
	school = {University of Colorado Boulder},
	author = {Burks, Charles Luke},
	month = may,
	year = {2020},
}

@article{ahmed_bayesian_2013,
	title = {Bayesian {Multicategorical} {Soft} {Data} {Fusion} for {Human}–{Robot} {Collaboration}},
	volume = {29},
	journal = {IEEE Transactions on Robotics},
	author = {Ahmed, Nisar R. and Sample, Eric M. and Campbell, Mark},
	month = feb,
	year = {2013},
	pages = {189--206},
}

@article{arora_survey_2021,
	title = {A survey of inverse reinforcement learning: {Challenges}, methods and progress},
	volume = {297},
	shorttitle = {A survey of inverse reinforcement learning},
	journal = {Artificial Intelligence},
	author = {Arora, Saurabh and Doshi, Prashant},
	month = aug,
	year = {2021},
	pages = {103500},
}

@inproceedings{sweet_structured_2016,
	address = {Boston, MA},
	title = {Structured synthesis and compression of semantic human sensor models for {Bayesian} estimation},
	booktitle = {2016 {American} {Control} {Conference} ({ACC})},
	publisher = {IEEE},
	author = {Sweet, Nicholas and Ahmed, Nisar},
	month = jul,
	year = {2016},
	pages = {5479--5485},
}

@article{tabrez_survey_2020,
	title = {A {Survey} of {Mental} {Modeling} {Techniques} in {Human}–{Robot} {Teaming}},
	volume = {1},
	journal = {Current Robotics Reports},
	author = {Tabrez, Aaquib and Luebbers, Matthew B. and Hayes, Bradley},
	month = dec,
	year = {2020},
	pages = {259--267},
}

@article{albrecht_autonomous_2018,
	title = {Autonomous {Agents} {Modelling} {Other} {Agents}: {A} {Comprehensive} {Survey} and {Open} {Problems}},
	volume = {258},
	shorttitle = {Autonomous {Agents} {Modelling} {Other} {Agents}},
	journal = {Artificial Intelligence},
	author = {Albrecht, Stefano V. and Stone, Peter},
	month = may,
	year = {2018},
	pages = {66--95},
}

@inproceedings{gervits_toward_2020,
	address = {Auckland, New Zealand},
	title = {Toward {Genuine} {Robot} {Teammates}: {Improving} {Human}-{Robot} {Team} {Performance} {Using} {Robot} {Shared} {Mental} {Models}},
	booktitle = {Proc. of the 19th {International} {Conference} on {Autonomous} {Agents} and {Multiagent} {Systems}},
	publisher = {IFAAMAS},
	author = {Gervits, Felix and Thurston, Dean and Thielstrom, Ravenna and Fong, Terry and Pham, Quinn and Scheutz, Matthias},
	month = may,
	year = {2020},
	pages = {429--437},
}

@article{scheutz_framework_2017,
	title = {A {Framework} for {Developing} and {Using} {Shared} {Mental} {Models} in {Human}-{Agent} {Teams}},
	volume = {11},
	number = {3},
	journal = {Journal of Cognitive Engineering and Decision Making},
	author = {Scheutz, Matthias and DeLoach, Scott A. and Adams, Julie A.},
	month = sep,
	year = {2017},
	pages = {203--224},
}

@inproceedings{ray_review_2022,
	address = {Dubrovnik, Croatia},
	title = {A {Review} of the {Operational} {Use} of {UAS} in {Public} {Safety} {Emergency} {Incidents}},
	publisher = {IEEE},
    booktitle = {Proc. of the International Conference on Unmanned Aircraft Systems {(ICUAS)}},
	author = {Ray, Hunter and Singer, Ryan and Ahmed, Nisar},
	year = {2022},
	pages = {922--931},
}

@inproceedings{miller_delegation_2014,
	address = {New York, NY},
	series = {{HCI}-{Aero} '14},
	title = {Delegation and intent expression for human-automation interaction},
	booktitle = {Proc. of the {International} {Conference} on {Human}-{Computer} {Interaction} in {Aerospace}},
	publisher = {Association for Computing Machinery},
	author = {Miller, Christopher A.},
	year = {2014},
}

@inproceedings{hadfield-menell_inverse_2017,
	title = {Inverse {Reward} {Design}},
	volume = {30},
	booktitle = {Advances in {Neural} {Information} {Processing} {Systems}},
	publisher = {Curran Associates, Inc.},
	author = {Hadfield-Menell, Dylan and Milli, Smitha and Abbeel, Pieter and Russell, Stuart J and Dragan, Anca},
	year = {2017},
}

@article{burks_harps_2022,
	title = {{HARPS}: {An} {Online} {POMDP} {Framework} for {Human}-{Assisted} {Robotic} {Planning} and {Sensing}},
	journal = {IEEE Transactions on Robotics},
	author = {Burks, Luke and Ray, Hunter M and McGinley, Jamison and Vunnam, Sousheel and Ahmed, Nisar},
	year = {2022},
    note = {In Press},
}

@article{arora_multi-modal_2019,
	title = {Multi-modal active perception for information gathering in science missions},
	volume = {43},
	journal = {Autonomous Robots},
	author = {Arora, Akash and Furlong, P. Michael and Fitch, Robert and Sukkarieh, Salah and Fong, Terrence},
	month = oct,
	year = {2019},
	pages = {1827--1853},
}

@article{rouse_looking_1986,
	title = {On looking into the black box: {Prospects} and limits in the search for mental models},
	volume = {100},
	copyright = {© 1986, American Psychological Association},
	shorttitle = {On looking into the black box},
	journal = {Psychological Bulletin},
	author = {Rouse, William B. and Morris, Nancy M.},
	month = nov,
	year = {1986},
	pages = {349--363},
}

@inproceedings{jamieson_active_2020,
	title = {Active {Reward} {Learning} for {Co}-{Robotic} {Vision} {Based} {Exploration} in {Bandwidth} {Limited} {Environments}},
	booktitle = {2020 {IEEE} {International} {Conference} on {Robotics} and {Automation} ({ICRA})},
	author = {Jamieson, Stewart and How, Jonathan P. and Girdhar, Yogesh},
	month = may,
	year = {2020},
	pages = {1806--1812},
}

@inproceedings{booth_perils_2023,
	address = {Washington, D.C},
	title = {The {Perils} of {Trial}-and-{Error} {Reward} {Design}: {Misdesign} through {Overfitting} and {Invalid} {Task} {Specifications}},
	booktitle = {Proc. of the 37th {AAAI} {Conference} on {Artificial} {Intelligence} ({AAAI})},
	author = {Booth, Serena and Knox, W Bradley and Shah, Julie and Niekum, Scott and Stone, Peter and Allievi, Alessandro},
	month = feb,
	year = {2023},
}

@article{jarvelin_cumulated_2002,
	title = {Cumulated gain-based evaluation of {IR} techniques},
	volume = {20},
	journal = {ACM Transactions on Information Systems},
	author = {Järvelin, Kalervo and Kekäläinen, Jaana},
	month = oct,
	year = {2002},
	pages = {422--446},
}

@book{bishop_pattern_2006,
	address = {Cambridge U.K.},
	title = {Pattern {Recognition} and {Machine} {Learning}},
	publisher = {Springer},
	author = {Bishop, Christopher M.},
	year = {2006},
}

@inproceedings{ahmed_fully_2015,
	address = {Seattle, Washington},
	title = {Fully bayesian learning and spatial reasoning with flexible human sensor networks},
	booktitle = {Proc. of the {ACM}/{IEEE} {Sixth} {International} {Conference} on {Cyber}-{Physical} {Systems} - {ICCPS} '15},
	publisher = {ACM Press},
	author = {Ahmed, Nisar and Campbell, Mark and Casbeer, David and Cao, Yongcan and Kingston, Derek},
	year = {2015},
	pages = {80--89},
}

@inproceedings{wakayama_active_2023,
	address = {London, UK},
	title = {Active {Inference} for {Autonomous} {Decision}-{Making} with {Contextual} {Multi}-{Armed} {Bandits}},
	author = {Wakayama, Shohei and Ahmed, Nisar},
	booktitle = {2023 {IEEE} {International} {Conference} on {Robotics} and {Automation}},
 year = 2023
}

@inproceedings{zheng2021multi,
  title={Multi-resolution {POMDP} planning for multi-object search in {3D}},
  author={Zheng, Kaiyu and Sung, Yoonchang and Konidaris, George and Tellex, Stefanie},
  booktitle={2021 IEEE/RSJ International Conference on Intelligent Robots and Systems (IROS)},
  pages={2022--2029},
  year={2021},
  organization={IEEE}
}

@inproceedings{wandzel2019multi,
  title={Multi-object search using object-oriented {POMDP}s},
  author={Wandzel, Arthur and Oh, Yoonseon and Fishman, Michael and Kumar, Nishanth and Wong, Lawson LS and Tellex, Stefanie},
  booktitle={2019 International Conference on Robotics and Automation (ICRA)},
  pages={7194--7200},
  year={2019},
  organization={IEEE}
}

@inproceedings{xiao2019online,
  title={Online planning for target object search in clutter under partial observability},
  author={Xiao, Yuchen and Katt, Sammie and ten Pas, Andreas and Chen, Shengjian and Amato, Christopher},
  booktitle={2019 International Conference on Robotics and Automation (ICRA)},
  pages={8241--8247},
  year={2019},
  organization={IEEE}
}

@inproceedings{baker2016planning,
  title={Planning search and rescue missions for {UAV} teams},
  author={Baker, Chris AB and Ramchurn, Sarvapali and Teacy, WT Luke and Jennings, Nicholas R},
  booktitle={Proceedings of the Twenty-second European Conference on Artificial Intelligence},
  pages={1777--1778},
  year={2016}
}

@inproceedings{kingston2016automated,
  title={Automated {UAV} tasks for search and surveillance},
  author={Kingston, Derek and Rasmussen, Steven and Humphrey, Laura},
  booktitle={2016 IEEE Conference on Control Applications (CCA)},
  pages={1--8},
  year={2016},
  organization={IEEE}
}

@article{thomas2013us,
  title={{US} coast guard addendum to the {US} {NSS} to the {IAMSAR}},
  author={Thomas, CB},
  journal={United States Coast Guard, Tech. Rep. COMDTINST MI6130},
  volume={2},
  year={2013}
}

@article{bourgault2006optimal,
  title={Optimal search for a lost target in a bayesian world},
  author={Bourgault, Fr{\'e}d{\'e}ric and Furukawa, Tomonari and Durrant-Whyte, Hugh F},
  journal={Field and Service Robotics: Recent Advances in Reserch and Applications},
  pages={209--222},
  year={2006},
  publisher={Springer}
}

@article{silver2010monte,
  title={Monte-Carlo planning in large {POMDPs}},
  author={Silver, David and Veness, Joel},
  journal={Advances in neural information processing systems},
  volume={23},
  year={2010}
}

@book{kochenderfer2022algorithms,
  title={Algorithms for decision making},
  author={Kochenderfer, Mykel J and Wheeler, Tim A and Wray, Kyle H},
  year={2022},
  publisher={MIT press}
}

@misc{ray2023humancentered,
      title={Human-Centered Autonomy for {UAS} Target Search}, 
      author={Hunter M. Ray and Zakariya Laouar and Zachary Sunberg and Nisar Ahmed},
      year={2023},
      eprint={2309.06395},
      archivePrefix={arXiv},
      primaryClass={cs.RO}
}

\appendix

\subsection{Model Inference for Reward Estimation}
The graphical model in Figure \ref{fig: graphical_model} shows us how to calculate the overall joint probability $p(r,\Psi,\Phi,\theta,\Delta,\mathcal{K},\mathcal{P},\mathcal{O},\mathcal{S})$. As we aim to approximate $P(s)$ through our reward proxy $r_g$ for all $g \in \mathcal{G}$, we therefore infer $\hat{r}$ as the expected value of $p(r|\Psi,\Phi,\theta,\Delta,\mathcal{K},\mathcal{P},\mathcal{O},\mathcal{S})$, which simplifies to $p(r,\theta,\Delta|\Psi,\Phi,\mathcal{P},\mathcal{O})$ based on conditional independence. A variety of methods can be used to perform inference via the graphical model, including Gibbs sampling \cite{ahmed_fully_2015}, variational Bayes \cite{ahmed_bayesian_2013}, or the Laplace approximation \cite{wakayama_active_2023}, among others \cite{bishop_pattern_2006}. 
From the chain rule, it follows
\begin{align}
    \label{eq:chainrule_post}
    &p(r,\theta,\Delta|\Psi,\Phi,\mathcal{K},\mathcal{P},\mathcal{O},\mathcal{S}) \nonumber \\ &=p(r|\theta,\Delta,\Psi,\Phi,\mathcal{K},\mathcal{P},\mathcal{O},\mathcal{S}) p(\theta,\Delta|\Psi,\Phi,\mathcal{K},\mathcal{P},\mathcal{O},\mathcal{S}) \nonumber \\
    &=p(r|\theta,\Delta,\Psi,\Phi) p(\theta,\Delta|\Psi,\Phi,\mathcal{P},\mathcal{S}),
\end{align}
where the second equality follows from the conditional independence properties of the graphical model. 
From this, the posterior expected reward $\mathbb{E}[r_{g_{i,j}}]$ can be approximated according to equation \ref{eq: reward_est} if only the joint posterior expected values and variances for $(\theta,\Delta)$ are considered via the second factor in the RHS of \ref{eq:chainrule_post}.   

Thus, we focus on approximating this second posterior factor $p(\theta,\Delta|\Phi,\Psi,\mathcal{P},\mathcal{S})$. We require the inference to be flexible to a changing feature set and simple enough to be run in real-time. The Laplace approximation is well suited for these requirements. 

From Bayes rule, the desired posterior factor is proportional to
\begin{align}
\label{eq: propto}
    p(\theta,\Delta|S,\Phi,\Psi,\mathcal{P}) \propto p(\theta,\Delta|\mathcal{P})p(S|\Phi,\Psi,\theta,\Delta). 
\end{align}
Since each $S_i$ is conditionally independent of the other, 
\begin{align}
\label{eq: independence}
    p(\theta,\Delta|\mathcal{P})p(S|\Phi,\Psi,\theta,\Delta)&= \nonumber\\ 
    p(\theta,\Delta|\mathcal{P})&\prod_{g=1}^K p(S_g|\Phi_g,\Psi_g,\theta,\Delta).
\end{align}

With $p(\theta,\Delta|\mathcal{P})$ modeled via Gaussian priors and $p(S_g|\Phi_g,\Psi_g,\theta,\Delta)$ modeled as logistic functions, the LHS of \ref{eq: propto} is not analytically tractable.
We leverage the Laplace approximation to approximate the RHS of (\ref{eq: independence}) and the normalizing constant $C$ for (\ref{eq: propto}), thus permitting a Gaussian approximation of $p(\theta,\Delta|S_{1\dots k},\Phi,\Psi,\mathcal{P})$. 
Specifically, the posterior is approximated by fitting a Gaussian distribution over $f(\theta,\Delta)$,
where the mean is equal to the MAP estimate ($\theta^*,\Delta^*$) (obtained via quasi-Newton optimization on $\log f(\theta,\Delta)$) and the covariance matrix is the inverse of the Hessian $\textbf{A} = \textbf{H}[\log f(\theta,\Delta)]$ \cite{bishop_pattern_2006}, i.e. 
\begin{align}
\label{eq: laplace func}
           f(\theta,\Delta) &= p(\theta,\Delta|\mathcal{P})p(S|\theta,\Delta,\Phi,\Psi) \cdot C \nonumber\\
      &\approx {\cal N}\left(
\begin{bmatrix}
\theta^*\\
\Delta^*
\end{bmatrix}
, \textbf{A}^{-1}\right).
\end{align}
The computational demand for inference is driven by inversion of $\textbf{A}$, giving worst case complexity of $O((n_{\Phi}+n_{\mathcal{O}})^3)$. The probabilistic model and inference approximations were implemented in Julia; calculations take $<1$ sec on a consumer laptop after compilation for the data presented next.

\subsection{Validation Metrics}
\begin{figure}[h]
    \centering
    \includegraphics[width=\linewidth]{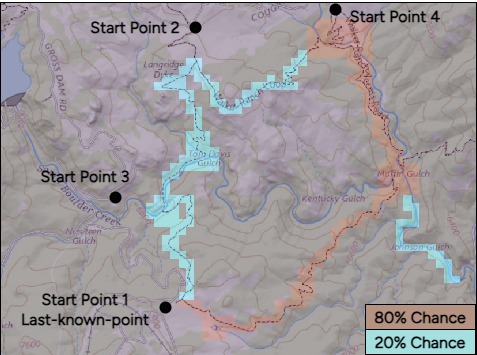}
    \caption{Expert informed sampled target locations with start points.}
    \label{fig: truth_sample}
    \vspace{-10pt}
\end{figure}
To model the total geospatial value error, subjects provided a specific reward value for a subset of 21 specific locations on a integer scale of -1 (Avoid) to 2 (Important to visit), whose more limited range simplifies the qualification task and reduces decision fatigue. The locations were hand selected to be geographically separated and to contain a diverse set of features.

Total error ($e$), aims to directly compare $\hat{r}_g$ with the user provided reward value, $r_{g,\textit{true}}$, of the 21 locations. As $r$ is modeled as a distribution, we weight the specific point estimate by $\textrm{var}(\hat{r}_{g_{i,j}})$ as defined in equation \ref{eq: reward_var} under the Laplace approximation, such that a more uncertain estimate of $\hat{r}_g$ has a lower impact on the total error. To match the operator's $\left\{-1,0,1,2\right\}$ ranking, we partition $\hat{r}$ across $\mathcal{G}$ by range into quartiles and assign a respective quantity of $\left\{-1,0,1,2\right\}$,
\begin{align}
\label{eq: error}
    e = \sum^{21}_g\frac{(\hat{r}_g-r_{g,\textit{true}})^2}{\mathrm{var}(\hat{r}_g)}.
\end{align}

To compare the realized total error against a randomized baseline, we applied equation \ref{eq: error} with a randomly selected $\hat{r}$ with a variance of 1.

Evaluating the ranking metrics required subject to define the relevance for each component of $\Phi$ and $\Psi$ on a Likert scale from -7 (Important to avoid) to 7 (Important to visit), which provides a wide range to account for nuanced preferences. This metric aims to find a set of preferences that captures the most valuable features to visit and those to avoid, since the absolute value of each set of weights isn't as important as its relative ranking between features.The adapated normalized Discounted Cumulative Reward (nDCG) \cite{jarvelin_cumulated_2002} metrics is compared against the subject's provided relevance, \textit{rel}$_i$, which is used to inform the Idealized Discounted Cumulative Gain (IDCG). IDCG ranks the set of available features based on the operator's set of \textit{rel}$_i$, similarly ranking in descending order for positive preferences and ascending order for negative preferences. The final Monte Carlo nDCG value (MCnDCG), considers a value of 1 to be optimal, 
\begin{align}
\label{eq: MCnDCG}
    \textrm{MCnDCG} = \sum^N_{s=1}\left[\sum^F_{i=1} \frac{\textit{rel}_i}{\log_2(i+1)}\frac{1}{IDCG} \right]\frac{1}{N}.
\end{align}
We found that 1000 Monte Carlo samples proved to create adequate results. For comparison, a set of $F$ values were uniformly sampled and ranked as a comparison.

\end{document}